# TF-MCL: Time-frequency Fusion and Multi-domain Cross-Loss for Self-supervised Depression Detection

Li-Xuan Zhao [a,§], Chen-Yang Xu [a,§], Wen-Qiang Li [a], Bo Wang [b,c], Rong-Xing Wei [b,c], Qing-Hao Meng [a,*]

[a]*School of Electrical and Information Engineering, Tianjin University, 300072, Tianjin, China*

[b]*Nanchang Police Dog Base of the Ministry of Public Security, Nanchang 330100, China*

[c]*Jiangxi Provincial Key Laboratory of Police Dog Breeding and Behavioral Science, Nanchang 330100, China*

**Abstract**

In recent years, there has been a notable increase in the use of supervised detection methods of major depressive disorder (MDD) based on electroencephalogram (EEG) signals. However, the process of labeling MDD remains challenging. As a self-supervised learning method, contrastive learning could address the shortcomings of supervised learning methods, which are unduly reliant on labels in the context of MDD detection. However, existing contrastive learning methods are not specifically designed to characterize the time-frequency distribution of EEG signals, and their capacity to acquire low-semantic data representations is still inadequate for MDD detection tasks. To address the problem of contrastive learning method, we propose a time-frequency fusion and multi-domain cross-loss (TF-MCL) model for MDD detection. TF-MCL generates time-frequency hybrid representations through the use of a fusion mapping head (FMH), which efficiently remaps time-frequency domain information to the fusion domain, and thus can effectively enhance the model's capacity to synthesize time-frequency information. Moreover, by optimizing the multi-domain cross-loss function, the distribution of the representations in the time-frequency domain and the fusion domain is reconstructed, thereby improving the model's capacity to acquire fusion representations. We evaluated the performance of our model on the publicly available datasets MODMA and PRED+CT and show a significant improvement in accuracy, outperforming the existing state-of-the-art (SOTA) method by 5.87% and 9.96%, respectively.

Key Words: MDD detection, Contrastive learning, Time-frequency fusion, Multi-domain cross-loss function

## 1. Introduction

Major depressive disorder (MDD) [1] is a prevalent emotional dysfunction that can manifest itself through a variety of physiological signals. Electroencephalogram (EEG) has been utilized to diagnose MDD and other psychiatric disorders due to its non-invasive, convenient, and efficient advantages. Presently, a considerable number of researchers employ the acquired EEG signals and integrate machine learning (ML) [2] or deep learning (DL) [3], [4] methods to distinguish MDD from healthy controls (HC). However, the general supervised ML or DL methods suffer from the limitation of over-reliance on data labeling. Consequently, it is frequently required that relevant psychologists conduct a large number of manual diagnoses in order to categorize the data, which requires the allocation of considerable medical resources.

Self-supervised methods could solve the problem of over-reliance on labeling in supervised methods. In recent years, the applications of contrastive learning methods in the field of time-series signals have gradually increased. Eldele et al. [5] proposed an unsupervised time-series representation learning framework via temporal and contextual contrasting (TSTCC), which designs a new cross-view prediction task to learn robust temporal representations. Yue et al. [6] put forward a universal framework for learning representations of time series in an arbitrary semantic level (TS2Vec), which implements a robust contextual representation for each timestamp by learning in a hierarchical contrastive manner in the augmented context view. Guo et al. [7] presented a modality consistency-guided contrastive learning (MoCL) method, which exploits the complementarity and redundancy between different time-series signals to construct a generalized model for personalized domain adaptation. Wu et al. [8] put forth an end-to-end auto-augmentation contrastive learning (AutoCL) method for time-series signals. AutoCL automatically learns data augmentation strategies, thereby alleviating the burden of manually designing such strategies.

---

\* Corresponding Author: Qing-Hao Meng, Email: qhmeng@tju.edu.cn. Li-Xuan Zhao and Chen-Yang Xu contributed equally.

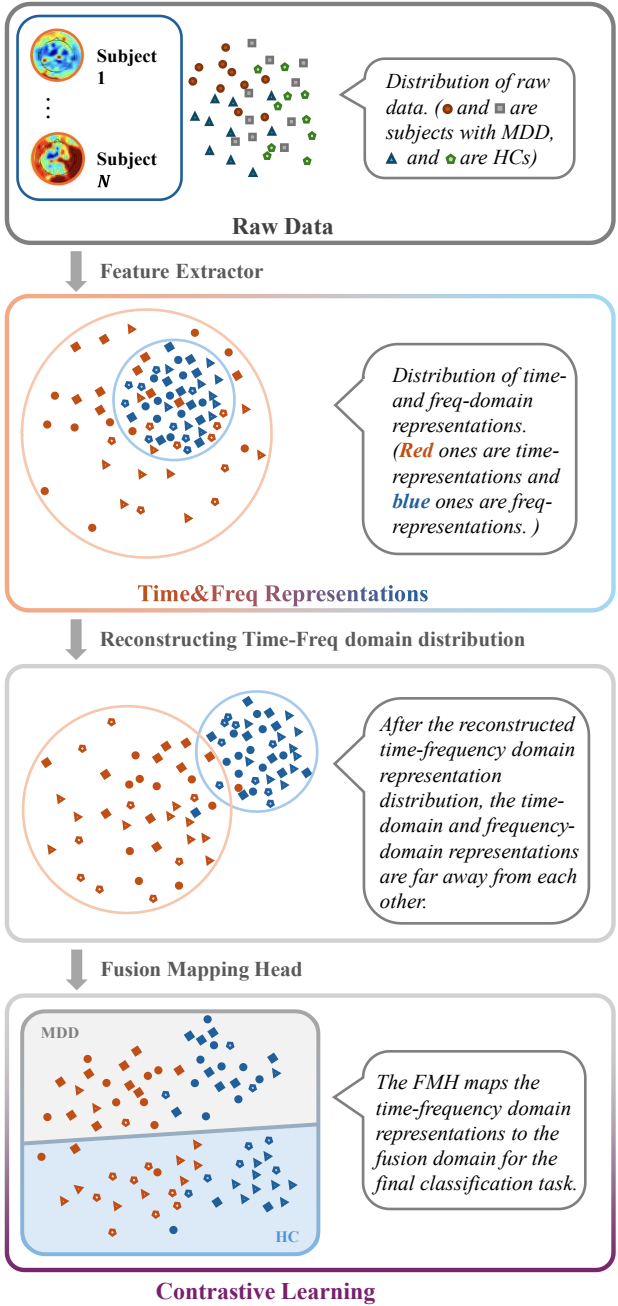

Figure 1: Distribution of data in TF-MCL.

consistency in the TFC [10] model elucidates the correlation between time-domain data and its frequency-domain transformation. Consequently, the effective synthesis of T-F information and generation of representations are of great significance for the improvement of model performance. As shown in Fig. 1, the spectral distribution of EEG exhibits minimal inter-individual variability, posing a challenge for models to effectively capture frequency domain information. Although TFC allows the T-F domain representations of the data to be close to each other in the potential space, it does not guarantee that the time-domain representations do not lose their diversity due to their proximity to the frequency-domain representations, which can lead to a degradation of model performance.

To address the aforementioned issues, we first propose fusion mapping head (FMH), a module that acts after the model feature extractor, which remaps T-F features to the fusion domain using self-attention (SA) and multi-layer perception (MLP) to effectively extract the correlation of T-F features. Second, we introduce a multi-domain cross-loss (MCL) function, which reconstructs the distribution of representations in the T-F and fusion domains. By optimizing the MCL function, the model extracts as many frequency-domain features as possible while ensuring that the time-domain representations do not lose their variability. Our contributions can be summarized as follows:

- For the T-F domain features of the original data, we propose FMH, a module that effectively integrates the features from the T-F domain of the samples and uniformly maps them to a potential fusion domain. The introduction of the FMH allows the model to further learn correlations with respect to the time-frequency domain features, which significantly improves the performance of the model.
- To fully utilize the T-F domain features of the original signals, the MCL is proposed. MCL effectively controls the separation and aggregation of sample representations in the time, frequency, and fusion domains from several perspectives, thereby generating finer-grained representations. This method solves the problem of missing diversity due to the proximity of T-F domain representations to each other. The results show that this method plays a very important role in improving the model performance.
- Our extensive experiments on two public datasets show that our method outperforms the existing state-of-the-art (SOTA) methods.

In the context of time-series signals, frequency domain information constitutes a pivotal feature. Consequently, scholars investigate contrastive learning methods based on time-frequency feature fusion. Yang et al. [9] introduced a new unsupervised time-series representation learning method called bilinear temporal-spectral fusion (BTSF), which can obtain excellent performance through a novel iterative bilinear time-frequency fusion method to explicitly model cross-domain dependencies. However, BTSF does not efficiently construct the respective time and frequency domain data for the samples and their enhanced versions. Zhang et al. [10] designed a time-frequency based consistent contrastive learning framework (TFC), which innovatively introduces the concept of time-frequency (hereafter T-F) consistency and enhances the generalization ability of contrastive learning methods. The concept of T-F

## 2. Related Works

Compared to supervised methods for the classification of depressive disorders, fewer studies have been conducted so far on self-supervised based methods, and even fewer on time-frequency fusion based contrastive learning methods in this field. Therefore, in order to introduce the research background of the proposed method in this paper, we not only summarize and analyze the studies related to depression recognition models using EEG, but also investigate contrastive learning methods for other physiological signals.

Duan et al. [11] extracted asymmetric and cross-correlated EEG features between the cerebral hemispheres and formed a hybrid feature by adding or subtracting these two features to represent brain states more comprehensively.

The study used three classifiers, K nearest neighbors (KNN), support vector machines (SVM) and convolutional neural networks (CNN), to validate the effectiveness of the features and achieved better results in depression detection tasks. Rafiei et al. [11] designed an effective channel selection strategy to reduce redundant channels and improve model efficiency, while using a customized InceptionTime model to achieve 91.67% accuracy on the test set. Xia et al. [13] automatically learned the connectivity relationship between EEG channels through a multi-head attention mechanism and used a parallel two-branch CNN to extract high-level features at different latitudes, which achieved better results on the publicly available dataset provided by Mumtaz et al. [14]. Wang et al. [15] designed a multi-task deep learning framework for MDD detection, which improved the overall performance of the model by contrasting the noise robustness task, the supervised feature extraction task, and the multi-task learning module. Uyulan et al. [16] tested the performance of ResNet-50, MobileNet, and Inception-v3 models for depression recognition, and found that MobileNet performs best at spatial resolutions of 89.33% for the left hemisphere and 92.66% for the right hemisphere. Wang et al. [17] designed DiffMDD based on the Diffusion and Transformer models to increase the diversity and quantity of data through a bidirectional diffusion noise training module, thus helping the model learn more generalized features.

Contrastive learning, as an effective self-supervised learning method, has shown great potential in the field of biomedical signal processing. By designing data augmentation methods and contrast loss functions for specific signals, contrastive learning is able to learn more robust and discriminative signal representations, thus improving the performance of the model in various tasks. In the future, contrastive learning is expected to be applied in more biomedical signal processing tasks, especially in multimodal signal fusion and weakly supervised learning, which will bring more innovations and breakthroughs in healthcare.

Cheng et al. [18] promoted subject-invariance by designing subject-specific contrastive loss and adversarial training. In addition, a series of time-series data augmentation techniques were developed in this study for training biological signals along with contrastive loss. Shen et al. [19] proposed the contrastive imagination modality sleep network (CIMSleepNet) framework, which aims to address the performance degradation of multimodal physiological signals in automated sleep staging (ASS) due to modality loss, through the modality aware imagination module (MAIM) and semantic & modal calibration contrastive learning (SMCCL) to deal with the problem of arbitrary modal absence, and multilevel cross-branching temporal attention mechanism (MCTA) was designed to mine cross-scale temporal contextual representations. Rabbani et al. [20] proposed a contrastive self-supervised learning (SSL)-based model for assessing pressure from ECG signals. The model is based on the SimCLR framework and utilizes unlabeled ECG data through contrastive learning. Matton et al. [21] proposed a method based on a contrastive learning framework for detecting stress from electrodermal activity (EDA) signals. The method utilizes unlabeled EDA data to learn a representation of stress assessment through self-supervised learning and achieves significant performance gains in downstream stress detection tasks. Sun et al. [22] presented a method called Contrast-Phys+ for video-based telephysiological measurements, especially in unsupervised and weakly-supervised environments, which is capable of measuring blood volume change signals (i.e., remote photovolumetric tracing, rPPG) via facial videos. The method demonstrates good performance and generalization ability by generating multiple spatiotemporal rPPG signals using 3D CNN models trained with prior knowledge and contrastive loss functions, which are able to learn efficiently with no or only partial labels.

## 3. Methodology

This section presents the TF-MCL framework, as shown in Fig. 2, which encompasses the input generation and data augmentation, T-F domain feature extractor $G = (G_T | G_F)$, FMH and MCL components. Given the unlabeled pre-training dataset $D^{pret} = \{x_i^{pret} | i = 1,2,\ldots,N^{pret}\}$ and the fine-tuning dataset $D^{tune} = \{(x_i^{tune}, y_i^{tune}) | i = 1,2,\ldots,N^{tune}\}$, where $x_i^{pret} \in D^{pret}$ is the unlabeled sample and $(x_i^{tune}, y_i^{tune}) \in D^{tune}$ denotes the labeled sample, $dim(x_i^{pret}) = dim(x_i^{tune}) = (E, T)$, where $E$ indicates the number of electrode channels, and $T$ stands for the number of timestamps corresponding to the sample slices. The objective is to pre-train $G$ and FMH with $D^{pret}$, and then fine-tune the parameters on $D^{tune}$, so that the fine-tuned model will have a good representation for each $x_i^{tune}$.

### 3.1 Time-frequency Input Generation

EEG signals have the typical characteristics of time-series signals, with features that respond to changes in the signal over time. However, the time-domain features cannot respond to the energy distribution characteristics of EEG signals. Therefore, the power spectral density (PSD) [23] of the original signal is introduced in TF-MCL as one of the inputs of the model. Based on the above analysis, frequency-domain features are essential for MDD detection. Therefore, we use the autocorrelation method to calculate the PSD. Given a single sample $x_i^{pret} \in D^{pret}$, we choose a single channel $E_p$, and noting that $x_{i,E_p} = x_i^{pret}[E_p]$, then $dim(x_{i,E_p}) = T$, and its autocorrelation function $R_{x_{i,E_p}}[k]$ is defined as:

$$R_{x_{i,E_p}}[k] = \lim_{N \to \infty} \frac{1}{2N+1} \sum_{n=0}^{2N} x_{i,E_p}[n] x_{i,E_p}[n+k], \quad (1)$$

where $x_{i,E_p}[n]$ denotes the time step $n$ of $x_{i,E_p}$. According to Eq. (1), $PSD(x_{i,E_p}, e^{j\omega})$ can be calculated as:

$$PSD(x_{i,E_p}, e^{j\omega}) = \sum_{k=-\infty}^{\infty} R_{x_{i,E_p}}[k] e^{-j\omega k}, \quad (2)$$

where $x_{i,E_p}$ is the input raw signal and $R_{x_{i,E_p}}[k]$ indicates its autocorrelation function [23]. Compared to the Fourier transform that responds to the band relationship, PSD responds to the relationship and distribution of the energy of the EEG channels, which helps to extract more features from low semantic EEG signals to improve the accuracy of MDD detection [24].

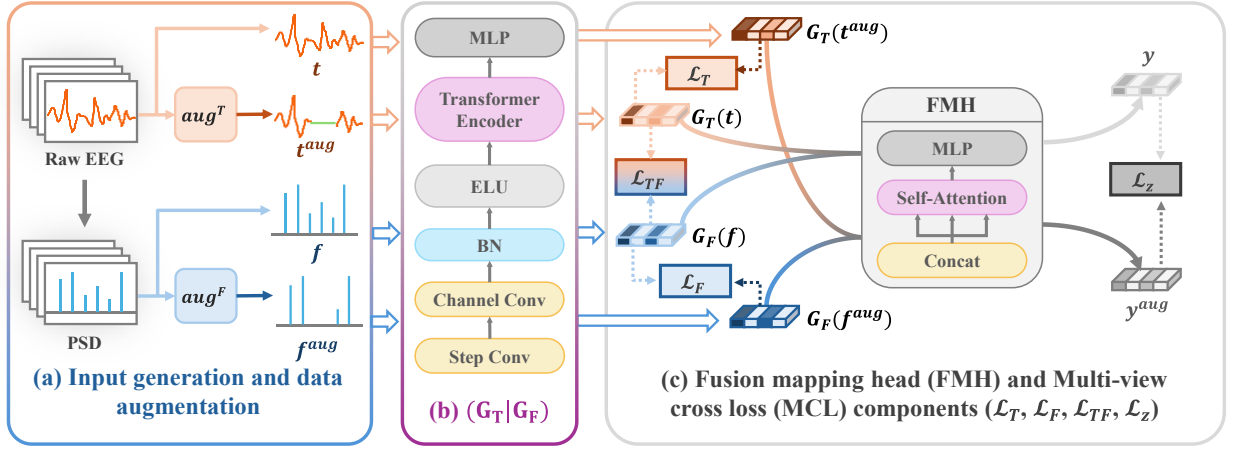

Figure 2: Framework of TF-MCL.

The generated T-F domain data are fed into the TF-MCL after data augmentations, which improves the generalization of the model and forces the model to learn the essential features of the data. The time-domain data augmentation methods selected for investigation are resampling, random channel substitution, timing inversion, noise addition, and channel perturbation. In addition, frequency-domain data augmentation methods, such as band removal and band addition, are employed [25].

We denote a batch size $N$ of EEG data and its corresponding PSD representation as $t$ and $f$, and the corresponding data after data augmentations are denoted as $t^{aug}$ and $f^{aug}$, respectively.

### 3.2 Multidomain Cross-Loss Components

MCL plays a pivotal role in the reconstruction of the multi-domain representation distribution throughout the process of model optimization. The feature extractor $G$ generates T-F domain representations $rt$, $rf$ about the original data and the respective augmented version $rt^{aug}$, $rf^{aug}$, i.e.

$$rt = G_T(t), rt^{aug} = G_T(t^{aug}),$$
$$rf = G_F(f), rf^{aug} = G_F(f^{aug}), \quad (3)$$

where $G_T(\cdot)$, $G_F(\cdot)$ denote the time-domain feature extractor as well as frequency domain feature extractor, respectively. They have different step-convolutional kernel sizes to fit the input data dimensions. $rt$, $rf$ indicate the corresponding representations, and $rt^{aug}$, $rf^{aug}$ are augmented versions. The feature extractor $G$ extracts features from the input data and its augmented version using efficient convolutional layers as well as a transformer encoder. As shown in Fig. 2 (c), the T-F representations are generated via FMH, i.e.

$$y = f(G_T(t) \oplus G_F(f)),$$

$$y^{aug} = f(G_T(t^{aug}) \oplus G_F(f^{aug})), \quad (4)$$

where $f(\cdot)$ denotes the mapping of the neural network FMH to the representation, and the notations $G_T(t) \oplus G_F(f)$ and $G_T(t^{aug}) \oplus G_F(f^{aug})$ indicate the splicing of the corresponding representations. The variables $y$ and $y^{aug}$ are generated after FMH, which synthesizes and reorganizes the T-F representations. This process ultimately generates effective fusion representations.

Let $Z = y \oplus y^{aug}$, i.e. $Z = \{z_i | z_i = y, i \leq N; z_i = y_i^{aug}, N < i \leq 2N\}$, and based on the training objective, the output representations should be close to each other, so it is intuitive to define the normalized temperature scale cross-entropy loss function (NT-Xent) [26] for individual samples and their augmented versions:

$$\ell(z_i, z_j) = -\log \frac{\exp(sim(z_i, z_j)/\tau)}{\sum_{k=1}^{2N} \mathbf{1}_{[k \neq i]} \exp(sim(z_i, z_j)/\tau)}, \quad (5)$$

where $N$ is the batch size, $\tau$ denotes a temperature hyperparameter which regulates the uniformity of the distribution, $sim(z_i, z_k)$ indicates the cosine similarity between representations, and $\mathbf{1}$ stands for the indicator function that equals 1 if $k \neq i$ and 0 otherwise. For the entire batch, we can define the fused representation loss (FRL) as follows:

$$\mathcal{L}_z = \frac{1}{2N} \sum_{k=1}^{N} [\ell(z_{2k-1}, z_{2k}) + \ell(z_{2k}, z_{2k-1})], \quad (6)$$

where $\mathcal{L}_z$ reflects how close the representations are in the fusion domain. As the primary component of MCL, FRL is responsible for extracting the essential T-F features of the original EEG signal.

According to Eq. (5) and Eq. (6), for the T-F representations $rt, rt^{aug}, rf, rf^{aug}$ generated by Eq. (3), the time-domain loss (TL) and frequency-domain loss (FL) can be expressed as the following loss functions:

$$\mathcal{L}_T = \frac{1}{2N} \sum_{k=1}^{N} [\ell(r_{2k-1}^T, r_{2k}^T) + \ell(r_{2k}^T, r_{2k-1}^T)], \quad (7)$$

$$\mathcal{L}_F = \frac{1}{2N}\sum_{k=1}^{N}[\ell(r_{2k-1}^F, r_{2k}^F) + \ell(r_{2k}^F, r_{2k-1}^F)], \quad (8)$$

where $\mathcal{L}_T$ and $\mathcal{L}_F$ denote TL and FL, respectively. $r_i^T$ equals $rt_i$ if $i \leq N$ and $rt_i^{aug}$ otherwise, and $r_i^F$ equals $rf_i$ if $i \leq N$ and $rf_i^{aug}$ otherwise. In both cases, $rt_i$, $rf_i$, $rt_i^{aug}$, $rf_i^{aug}$ denote the input representations. TL and FL are employed to quantify the degree of similarity between the original data and its augmented version, which enhances the model's capacity to extract representations from both the time and frequency domains.

Furthermore, to guarantee the variability of the T-F domain representations and thereby obtain more information and ensure optimal model performance, we define T-F dispersion loss (TFDL) $\mathcal{L}_{TF}$ as follows:

$$\mathcal{L}_{TF} = \left(\frac{1}{N}\sum_{i=1}^{N} \exp\left(\frac{rt_i \cdot rf_i}{||rt_i||\,||rf_i||}\right)\right)^{-1}, \quad (9)$$

where $\mathcal{L}_{TF}$ plays a pivotal role in shaping the learning trajectory of the model by regulating the distance between the representations in the T-F domain. This enables the model to discern more intricate features in the T-F domain data. Therefore, it is sufficient to employ solely the original version of the time-frequency domain representations, $rt_i$ and $rf_i$, for control purposes. The utilisation of the enhanced versions of $rt^{aug}$ and $rf^{aug}$ will result in the control of $\mathcal{L}_{TF}$ over the learning direction of the model being weaker, due to the effect of the data enhancement, as will be demonstrated subsequently in the experiments. Where $||rt_i||$ denotes the L2-norm of the representation. From Eq. (9), $\mathcal{L}_{TF}$ calculates the cosine similarity between the representations corresponding to the timing signals as well as the frequency signal outputs of a single slice of EEG data. The smaller the value of TFDL, the more far away from each other the T-F representations are in the projected space.

The combination of TFDL and FRL enables the models to extract their distinctive features in the time and frequency domains, while ensuring that the output representations are similar in the fusion domain. Furthermore, the integration of TL and FL enhances the model's capacity to extract T-F features, thereby improving the model's performance in the time, frequency, and fusion domains.

*3.3 Loss Function*

We combine the above-mentioned time and frequency domain losses (TL and FL), fusion representation loss (FRL), T-F dispersion loss (TFDL), and the final multi-domain crossover loss (MCL) as follows.

$$\mathcal{L} = \alpha(\mathcal{L}_T + \mathcal{L}_F) + (1-\alpha)\mathcal{L}_z + \beta\mathcal{L}_{TF}. \quad (10)$$

MCL consists of FRL and TFDL, where $\alpha$ and $\beta$ are the corresponding influence coefficients, which can be adjusted to improve the model performance. Where $\alpha$ indicates the importance of fusion domain representation aggregation compared to T-F domain aggregation, and $\beta$ denotes the influence of T-F dispersion in the fusion cross-comparison loss function.

*3.4 Baseline Network*

In this study, we utilize TFC [10] as the baseline network and its initial version of this network adopts ResNet [27] as the T-F feature extractor. Based on the multichannel and low semantic characteristics of EEG signals, we design a simple and effective feature extractor $G$, as illustrated in Fig. 2 (b). The single slice of raw EEG signals contains a substantial amount of data, and the direct application of SA will result in a prolonged inference time. Therefore, the feature extractor initially employs time/frequency step convolution, then the channel convolution is used to compress and integrate the data, after which the transformer encoder [28] based on the single-head attention mechanism and the single-layer feed-forward network is utilized to conduct further feature extraction.

## 4. Experiments

*4.1 Dataset Description and Preprocessing*

- **Datasets** Extensive experiments were conducted on two public datasets, MODMA [29] and PRED+CT [30]. Unnecessary electrodes (CB1, CB2, HEOG, VEOG, EKG) were removed, and TP9 and TP10 in PRED+CT were re-referenced.
- **Processing** In the data preprocessing stage, a band-pass filter in the range of 0.1 Hz to 40 Hz was employed to eliminate low- and high-frequency noise, while a trap filter in the range of 49 Hz to 51 Hz was utilized to attenuate industrial frequency interference. Subsequently, we employed independent component analysis (ICA) to eliminate any artifactual components, such as those resulting from eye or head movements.

*4.2 Hyperparameters Setting*

The overall structure of the network is shown in Fig. 2. Table 1 summarizes the hyperparameter values used. Default values were used for the other hyperparameters.

Table 1: Hyperparameters Setting.

| Hyperparameters | MODMA | PRED+CT |
|---|---|---|
| Temporal Conv Kernel | (1,20) | (1,10) |
| Frequent Conv Kernel | (1,10) | (1,5) |
| Channel Conv Kernel | (60,1) | (128,1) |
| The head of SA in $G$ | 1 | 1 |
| Transformer Encoder Layers | 1 | 1 |
| The FC Layer of MLP in $G$ | 1 | 1 |
| The head of SA in FMH | 1 | 1 |
| The FC Layer of MLP in FMH | 3 | 3 |
| Optimizer | Adam | Adam |
| Epoch of pre-training | 100 | 100 |
| Epoch of fine-tuning | 200 | 200 |
| Batch Size | 128 | 128 |

Table 2: Results on Public Datasets (%).

| Method | Acc&F1 (MODMA) | Acc&F1 (PRED+CT) |
|---|---|---|
| TFC-Transformer [10] | 75.69 & 80.16 | 79.29 & 86.10 |
| TFC-ResNet [10] | 73.33 & 81.11 | 72.71 & 80.96 |
| TSConv [31] | 72.44 & 72.81 | 70.75 & 75.10 |
| EEGNet [32] | 66.80 & 66.76 | 57.80 & 61.08 |
| SimCLR-ViT [34] | 75.64 & 80.95 | 75.22 & 80.34 |
| SimCLR-ResNet [26] | 71.78 & 79.56 | 66.47 & 73.63 |
| Ts2Vec [6] | 77.20 & 76.67 | 72.90 & 81.13 |
| TSTCC [5] | 75.64 & 82.71 | 78.78 & 85.77 |
| TNC [33] | 74.13 & 81.21 | 77.61 & 83.03 |
| **Ours** | **83.07 & 82.83** | **89.25 & 90.46** |
| ΔSOTA | 5.87↑ & 1.72↑ | 9.96↑ & 4.36↑ |

*4.3 Platform*

The models were trained and evaluated on a computer with a single NVIDIA RTX 3090 24 GB GPU, Intel E5-2678 CPU, and 64 GB memory. The tests were carried out using the Pytorch deep learning framework.

*4.4 Comparison Experiment*

In our study, we conducted a comprehensive comparison with previous research results to evaluate the effectiveness of our proposed method in the context of MDD detection. Despite the limited number of contrastive learning methods specifically designed for MDD detection, we identified and replicated several SOTA contrastive learning methods that have been successfully applied in the broader field of time-series signal processing. These methods include TFC [10], SimCLR [26], TSConv [31], EEGNet [32], TS2Vec [6], TSTCC [5], and TNC [33]. By leveraging these established techniques, we aimed to provide a robust benchmark for assessing the performance of our novel approach.

The experimental results, as detailed in Table 2, demonstrate the significant improvements achieved by our method over existing SOTA methods. Specifically, on the MODMA dataset, our method outperformed the SOTA methods in terms of both accuracy and F1 score. The improvements were substantial, with an increase of 9.96% in accuracy and 4.36% in F1 score. Similarly, on the PRED+CT dataset, our method also exhibited enhanced performance compared to the SOTA method, achieving improvements of 5.87% in accuracy and 1.72% in F1 score. These results highlight the superior efficacy of our approach in detecting MDD from time-series signals.

The key innovation of our approach lies in the novel application of the concept of multi-domain fusion. Unlike traditional methods that focus solely on temporal or frequency information, our method integrates both time and frequency (T-F) information in a unified framework. This multi-domain fusion strategy significantly enhances the model's ability to extract relevant features from complex time-series signals, thereby improving its overall performance. By effectively leveraging the complementary nature of time and frequency information, our model is better equipped to capture the nuanced patterns associated with MDD, leading to the observed improvements in accuracy and F1 score.

In summary, our study provides compelling evidence that the integration of multi-domain fusion in contrastive learning offers a powerful approach for MDD detection. The experimental results on both the MODMA and PRED+CT datasets underscore the potential of our method to advance the field of MDD detection by improving diagnostic accuracy and reliability. Future work will focus on further optimizing the multi-domain fusion strategy and exploring its application in other related domains.

*4.5 Ablation Study*

Ablation experiments were conducted on the TF-MCL model, and the results are shown in Table 3. The results indicate that the TF-MCL model achieved the most highly improved performance compared to the baseline model when FMH is combined with MCL: by 7.16% for the MODMA dataset and by 18.23% for the PRED+CT dataset. Accuracy improvements of 4.05% (MODMA) and 11.69% (PRED+CT) were achieved when both FMH and TFDL were added to the baseline model, and 0.8% (MODMA) and 0.16% (PRED+CT) when only TFDL was added to the loss function, which means that TFDL plays a negligible role when acting alone, and note that as mentioned above, MCL consists of FRL and TFDL, and FRL depends on FMH.

In conjunction with the previous discussion, the various parts of the MCL play a vital role in model performance improvement. Based on the data in Table 3, it can be found that TFDL acting alone has a very limited effect on the performance enhancement for both the MODMA dataset and the PRED+CT dataset. On the contrary, FRL, when acting alone, has a significant performance enhancement for the model. The above results not only indicate that the samples' representations in the fusion domain contain more feature information, but also justify the design of FMH with FRL. In addition, although the performance enhancement of the model is limited when TFDL acts alone, the combination of FRL can further reorganize the distribution of the representations in each domain, further forcing the model to extract more feature information, which improves the performance as well as the generalization of the model.

Table 3: Ablation Studies (%).

| Dataset | Baseline | +FRL | +TFDL | Acc | F1 |
|---|---|---|---|---|---|
| MODMA | √ | - | - | 75.91 | 75.26 |
|  | √ | √ | - | 79.96 | 78.88 |
|  | √ | - | √ | 76.71 | 76.67 |
|  | √ | √ | √ | **83.07** | **82.83** |
| PRED+CT | √ | - | - | 71.02 | 78.01 |
|  | √ | √ | - | 82.71 | 88.98 |
|  | √ | - | √ | 71.18 | 81.78 |
|  | √ | √ | √ | **89.25** | **90.46** |

### 4.6 Parameter Sensitivity Analysis

We analyzed the weights of FRL as well as TFDL in MCL, and tested the effect of the parameters $\alpha$ and $\beta$ in Eq. (10) on the model performance. For $\alpha$, we conducted an experimental analysis with $\beta = 0$ to test the effect of FRL on the model. According to Table 4, the MODMA and the PRED+CT datasets both have the best performance at $\alpha = 0.2$. A reduction in the value of $\alpha$ results in an increase in FRL and enhances the model performance. Conversely, as the percentage of FRL declines, the model shifts its focus from the aggregation of representations in the time and frequency domains to that in the fusion domain. This shift leads to a loss of correlation between time-frequency representations and a subsequent decline in model accuracy.

For the parameter $\beta$, we fixed the value of $\alpha$ for the best model performance at 0.2, so as to adjust the value of $\beta$ in order to test the extent to which the magnitude of TFDL affects the model performance. According to Table 5, the MODMA dataset has the best performance at $\alpha = 0.2, \beta = 1$, while the PRED+CT dataset has the best performance at $\alpha = 0.2$, $\beta = 2$. The model demonstrates the optimal performance when the $\beta$ value is relatively low. However, when $\beta$ is excessively high, the model prioritizes the distinction between time- and frequency-domain representations, leading to the extraction of superfluous information and a subsequent decline in accuracy.

Table 4: Performance variation when the weight $\alpha$ of FRL in Eq. (6) takes different values (%).

| $\alpha$ | MODMA | | PRED+CT | |
|---|---|---|---|---|
|  | Acc | F1 | Acc | F1 |
| 0.0 | 79.87 | 79.91 | 78.82 | 86.55 |
| 0.1 | 79.82 | **79.88** | 78.86 | 86.58 |
| 0.2 | **79.96** | 78.88 | **82.71** | **88.98** |
| 0.3 | 78.93 | 78.18 | 74.78 | 83.71 |
| 0.4 | 78.98 | 78.21 | 74.90 | 83.78 |
| 0.5 | 76.44 | 76.02 | 76.00 | 84.31 |
| 0.6 | 75.42 | 75.33 | 74.98 | 83.82 |
| 0.7 | 75.29 | 75.25 | 74.71 | 83.66 |
| 0.8 | 75.60 | 75.48 | 73.92 | 83.26 |
| 0.9 | 75.16 | 75.10 | 74.04 | 83.27 |
| 1.0 | 75.91 | 75.68 | 71.02 | 81.70 |

Table 5: Performance variation when the weight $\beta$ of FRL in Eq. (10) takes different values.

| MODMA | | | PRED+CT | | |
|---|---|---|---|---|---|
| $\beta$ | Acc | F1 | $\beta$ | Acc | F1 |
| 0 | 79.96 | 78.88 | 0 | 82.71 | 88.98 |
| 0.1 | 78.89 | 78.02 | 0.1 | 83.06 | 89.24 |
| 0.5 | **81.82** | **81.48** | 0.5 | 83.33 | 88.68 |
| 1 | 83.07 | 82.83 | 1 | 85.61 | 89.20 |
| 2 | 80.98 | 81.03 | 2 | **89.25** | **90.46** |
| 3 | 80.93 | 81.00 | 3 | 78.90 | 86.14 |
| 4 | 79.82 | 79.88 | 4 | 77.06 | 85.76 |

($\alpha = 0.2$)

The results of the above pair of experiments reinforce the point we have elaborated in our ablation experiments that the two components, FRL and TFDL, need to have an exact fit share to have an optimal positive effect in the learning process of the model. In addition, we found that when the TFDL share is too large, the negative effect on the model performance is devastating. Then, when FRL is over-represented, the negative effect of FRL is much smaller than that of TFDL, although it causes a decrease in the model performance. The above phenomenon arises due to the fact that TFDL only normalizes the distribution between time-and frequency-domain representations, and does not have a direct effect on the feature extraction process of the model, while FRL directly acts on the feature extraction process of the model in the fusion domain, which in turn directly affects the model's performance improvement of the model. The rationality of this structure design is to improve the feature extraction ability of the model in multiple dimensions, and at the same time improve the robustness of the model among different datasets.

### 4.7 Visualization Analysis

In the field of machine learning and data classification, visualizing the performance of models is crucial for understanding their strengths and weaknesses. According to existing literature [35] and [36], we have employed the confusion matrix as a powerful tool to visualize our classification performance. The confusion matrix is a table that is often used to describe the performance of a classification model on a set of test data for which the true values are known.

Fig. 4 presents a detailed comparison of the confusion matrices between the baseline model and the proposed model when applied to the MODMA and PRED+CT datasets. For the MODMA dataset, the proposed model shows a significant improvement in classification accuracy. Specifically, the error rate for the HC group decreases by 5.00%, while the error rate for the MDD group decreases by 11.46%. This indicates that the proposed model is more accurate in distinguishing between HC and MDD cases compared to the baseline model. When it comes to the PRED+CT dataset, the improvements are even more pronounced. The error rate for the HC group decreases by 8.92%, and the error rate for the MDD group decreases by a remarkable 32.12%. These substantial reductions in error rates highlight the superior performance of the proposed model in handling the complexities of the PRED+CT dataset.

The proposed method not only reduces the overall error rates but also narrows the gap in accuracy between the HC and MDD groups. This reduction in discrepancy suggests that the model is more balanced and reliable in its classification across different categories. The enhanced stability in performance is a key advantage, as it ensures consistent results across various datasets and scenarios.

Two key components of the proposed model contribute to these improvements: FMH and MCL. FMH facilitates the learning of correlations between T-F information by effectively fusing T-F domain features. This fusion process allows the model to capture more nuanced and comprehensive patterns in the data, leading to better classification performance.

On the other hand, MCL prompts the model to extract a greater number of features based on the diverse attributes of T-F domain data. By leveraging multiple channels of information, the model can better understand the underlying structure and variability of the data. This results in enhanced accuracy and robustness, making the model more resilient to noise and outliers in the datasets.

Figure 3: Confusion Matrixes.

## 5. Discussion

### 5.1 Is it necessary to introduce frequency domain transformations of raw data in the field of MDD detection?

Time-frequency analysis plays a pivotal role in EEG analysis. In contrast to time-domain analysis, frequency-domain variability offers a more nuanced and information-rich perspective, unlocking deeper insights into the intricate patterns of EEG signals. This study innovatively bridges the gap between time and frequency domains by extracting PSD from raw time-domain EEG signals. PSD encapsulates a wealth of statistical metrics that reflect the inherent characteristics of EEG signals, such as mean, variance, and higher-order moments, which collectively paint a comprehensive picture of the signal's behavior. By feeding this PSD-enhanced, multi-dimensional information into the feature extractor, the model gains access to a broader and more detailed feature space.

Table 6: Results of single Time-Domain/Freq-Domain inputs.

| **Time-Domain Inputs** | Acc (%) | F1 (%) |
|---|---|---|
| MODMA | 72.22 | 80.58 |
| PRED+CT | 78.86 | 84.06 |
| **Freq-Domain Inputs** | Acc | F1 |
| MODMA | 60.36 | 60.66 |
| PRED+CT | 58.12 | 66.55 |
| **TF-Domain Inputs** | Acc | F1 |
| MODMA | 83.07 | 82.83 |
| PRED+CT | 89.25 | 90.46 |

As evidenced in Table 6, the model's performance sees a marked improvement when integrated time-frequency signals are introduced, compared to models relying solely on single time-domain or frequency-domain inputs. This performance gain stems from the enriched feature representation enabled by the fusion of time-frequency information. The integrated approach allows the model to capture both the temporal dynamics and frequency-specific characteristics of EEG signals. For instance, in the time domain, the model can track the evolution of brain activity over time, such as the onset and offset of specific brain wave patterns. In the frequency domain, it can identify dominant frequency bands and their corresponding power distributions, which are often linked to distinct cognitive or emotional states. This dual-domain perspective enables the model to uncover hidden correlations and interactions between different aspects of EEG signals that might be missed when analyzing a single domain in isolation.

*5.2  Why is the T-F fusion so important in TF-MCL?*

In the realm of processing input time series, the contrastive learning model is capable of generating time-based and frequency-based representations. However, this approach falls short by overlooking the inter-correlation between these two types of representations. To address this limitation and create finer-grained representations of the original data, we have developed a T-F fusion representation module.

The primary function of this module is to integrate features in the T-F domain. It does so by stitching together T-F domain inputs and uniformly mapping them to a fusion domain. This innovative module takes into account the fusion consistency of time and frequency domain signals for the first time. Within the module, an internal attention mechanism comes into play. It combines the common features of both domains and searches for their inter-correlations. By doing this, the module is able to obtain a more comprehensive representation of the original data semantics. This representation maintains a closer relationship with the original data.

As demonstrated by the experimental results presented above, incorporating the T-F fusion characterization module brings about a noticeable improvement in model performance. The module enhances the model's ability to capture the complex patterns and inherent relationships within the data. This leads to more accurate and meaningful representations, which in turn can improve the effectiveness of downstream tasks such as classification or prediction.

*5.3  Why is T-F dispersion critical for improving model accuracy?*

Different human physiological signals have different frequency domain distribution characteristics. While the frequency domain distribution of EEG signals has a very similar paradigm in different people and the same person under different brain activities, the EEG signals of the same channels of two people were randomly selected from MODMA and PRED+CT datasets to be plotted by PSD, as shown in Fig. 4 below. In terms of potential spatial distributions in the field of MDD detection, time-domain characterization does not coincide with frequency-domain characterization. Instead, we believe that adding time-frequency scatter to the already aggregated time-frequency fusion representations and moving the time- and frequency-domain representations away from each other, while keeping the fusion representations closer to each other in the fusion domain, can enable the feature extractor to obtain more dimensional information, which can be integrated to improve the model performance. According to the above experimental results, the performance of the model with the addition of time-frequency scatter loss is further improved compared to the previous one.

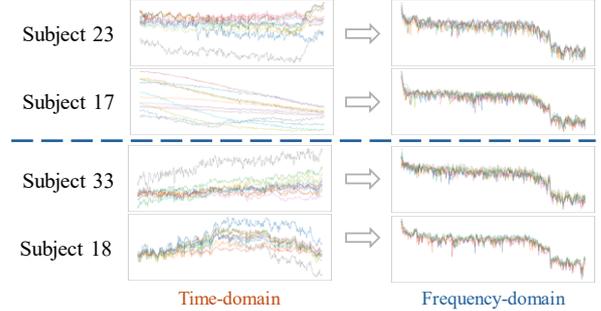

Figure 4: Power spectral density of EEG (Subjects 23 and 17 are patients with MDD, Subjects 33 and 18 are HC subjects, all randomly selected from the MODMA dataset.

*5.4  Why introduce $\mathcal{L}_{TF}$?*

In TF-MCL, the path of model convergence is to synthesize feature information about the samples by uniformly mapping the learned fine-grained features to the fusion domain for integration. In addition, it guarantees that as many features of the original data as possible can be learned from both the time-domain and frequency-domain data. In order to ensure that the model can learn different information from time-domain and frequency-domain representations to enhance its generalization performance, it is necessary to introduce $\mathcal{L}_{TF}$ in the MCL to regulate the convergence direction of the model. $\mathcal{L}_{TF}$ avoids the model learning convergent T-F domain representations in the T-F domain by regulating the cosine similarity between time-domain and frequency-domain representations, so that the distribution of the time-frequency domain representations in the latent domain is as far away as possible from each other, thus improves the feature extraction ability as well as the generalization of the model.

Regarding the representations of the data, there are original versions of $rt$, $rf$ and augmented versions of $rt^{aug}$, $rf^{aug}$. In $\mathcal{L}_{TF}$, we finally decided to use only the original versions for the computation. In order to improve the robustness and generalization of the model, the augmented version of the data introduces interfering information such as channel transformations and time transformations. This interfering information weakens the control of $\mathcal{L}_{TF}$ over the distribution of representations in the latent domain, which will degrade the performance of the model. The experimental results of introducing only the original version of the representations, only the augmented version and both in the $\mathcal{L}_{TF}$ are shown in Table 7. The results show that introducing only the original version of the representation provides the greatest improvement in model performance.

Table 7: Impact of raw & augmented data on the effectiveness of model feature extraction (%).

| Dataset | original | augmented | Acc | F1 |
|---|---|---|---|---|
| MODMA | √ | - | **83.07** | **82.83** |
|  | - | √ | 81.64 | 81.37 |
|  | √ | √ | 82.18 | 81.91 |
| PRED+CT | √ | - | **89.25** | **90.46** |
|  | - | √ | 86.31 | 87.65 |
|  | √ | √ | 87.61 | 88.89 |

## 6. Conclusion

We propose the first time-frequency fusion contrastive learning method, TF-MCL, in the field of MDD detection, which leads to the realization of a less-labeled, high-precision based detection model. Based on the baseline model, we design a fusion domain feature extractor with a simple but effective structure. In order to synthesize the time- and frequency-based characterization, we propose a time-frequency fusion characterization module that integrates and maps the extracted time-frequency features and optimizes them in conjunction with the time-frequency fusion characterization aggregation to remap the input data and integrate their distributions in the fusion domain to improve the model performance. Meanwhile, considering the distribution characteristics of EEG signals, we innovatively propose the concept of time-frequency scattering, and this module further enhances the ability of the model to access time-frequency information by differentiating the distribution of the time-frequency domain representations in the latent projection space, which improves the model accuracy for MDD detection. TF-MCL achieves high accuracy in MDD detection, outperforming the SOTA method by 5.87% and 9.96% in the MODMA and PRED+CT datasets, respectively.

We will continue to explore broader applications of contrastive learning methods in the field of physiological signaling. In the future, we will apply the contrastive learning model in multimodal physiological signal analysis and recognition, so as to make contributions to the field of human health.